\documentclass[accepted,specialissue]{melba}

\usepackage{amsmath,amsfonts}

\usepackage{multirow}
\usepackage{booktabs}
\usepackage{siunitx}
\usepackage{subcaption}

\usepackage{xcolor}
\usepackage{tikz}
\usepackage{tcolorbox}
\usepackage{todonotes}

\newcommand{\meanstd}[2]{#1$_{\pm\text{#2}}$}

\DeclareRobustCommand{\inlinering}[2]{%
  \tikz[baseline=-0.6ex]{%
    \draw[line width=#2, color=#1, fill=none] (0,0) circle(0.6ex);
  }%
}
\DeclareRobustCommand{\inlinecross}[3]{%
  \tikz[baseline=-0.6ex, x=#3, y=#3]{%
    \draw[line width=#2, color=#1] (-0.5,-0.5) -- (0.5,0.5);
    \draw[line width=#2, color=#1] (-0.5,0.5) -- (0.5,-0.5);
  }%
}
\definecolor{barbiepink}{RGB}{250, 140, 230}
\newcommand{\textexp}[2]{\textnormal{#1\textsuperscript{#2}}}

\melbaid{2026:004}  
\doi{10.59275/j.melba.2026-38ba}
\melbaauthors{Kahrs, Andresen, Falta, Santarossa, Handels, Kepp}  
\email{timo.kepp@dfki.de}
\volume{2026}
\firstpageno{59}  
\melbayear{2026}  
\datesubmitted{2025-07-15}  
\datepublished{2026-03}  

\melbaspecialissue{MELBA–BVM 2025 Special Issue}
\melbaspecialissueeditors{Andreas Maier, Thomas Deserno, Heinz Handels, Klaus Maier-Hein, Christoph Palm, Thomas Tolxdorff, Katharina Breininger}

\ShortHeadings{Resolution-Agnostic Retinal OCT Analysis}{Kahrs et al.}

\title{Don't Mind the Gaps: Implicit Neural Representations for Resolution-Agnostic Retinal OCT Analysis}

\author{
	\firstname Bennet \surname Kahrs\aff{1,2, *}\orcid{0009-0009-5609-9250},
	\firstname Julia \surname Andresen\aff{2,*}\orcid{0000-0002-9113-3954},
    \firstname Fenja \surname Falta\aff{2}\orcid{0009-0003-3234-9725},
    \firstname Monty \surname Santarossa\aff{3}\orcid{0000-0002-4159-1367},
    \firstname Heinz \surname Handels\aff{1,2}\orcid{0000-0002-3499-4328},
    \firstname Timo \surname Kepp\aff{1}\orcid{0000-0003-2024-2958}
}
\affiliations{
	\num 1 \addr German Research Center for Artificial Intelligence, Luebeck, DE \\
	\num 2 \addr Institute of Medical Informatics, University of Luebeck, Luebeck, DE \\
	\num 3 \addr Multimedia Information Processing Group, Kiel University, Kiel, DE \\
    \num * \addr These authors contributed equally to this work.
}
\abstract{
    Routine clinical imaging of the retina using optical coherence tomography (OCT) is performed with large slice spacing, resulting in highly anisotropic images and a sparsely scanned retina. Most learning-based methods circumvent the problems arising from the anisotropy by using 2D approaches rather than performing volumetric analyses. These approaches inherently bear the risk of generating inconsistent results for neighboring B-scans. For example, 2D retinal layer segmentations can have irregular surfaces in 3D. Furthermore, the typically used convolutional neural networks are bound to the resolution of the training data, which prevents their usage for images acquired with a different imaging protocol. Implicit neural representations (INRs) have recently emerged as a tool to store voxelized data as a continuous representation. Using coordinates as input, INRs are resolution-agnostic, which allows them to be applied to anisotropic data. In this paper, we propose two frameworks that make use of this characteristic of INRs for dense 3D analyses of retinal OCT volumes. 1) We perform inter-B-scan interpolation by incorporating additional information from en-face modalities, that help retain relevant structures between B-scans. 2) We create a resolution-agnostic retinal atlas that enables general analysis without strict requirements for the data. Both methods leverage generalizable INRs, improving retinal shape representation through population-based training and allowing predictions for unseen cases. Our resolution-independent frameworks facilitate the analysis of OCT images with large B-scan distances, opening up possibilities for the volumetric evaluation of retinal structures and pathologies.\\
    Our code is available at~\url{https://github.com/tkepp/ResA-OCT}.}%

\keywords{Optical Coherence Tomography, Implicit Neural Representation, B-scan Interpolation, Retinal Atlas, Multi-modal Analysis, Image Registration}

\begin{document}

\twocolumn[\maketitle]

\section{Introduction}

\begin{figure*}[t]
\includegraphics[width=\textwidth]{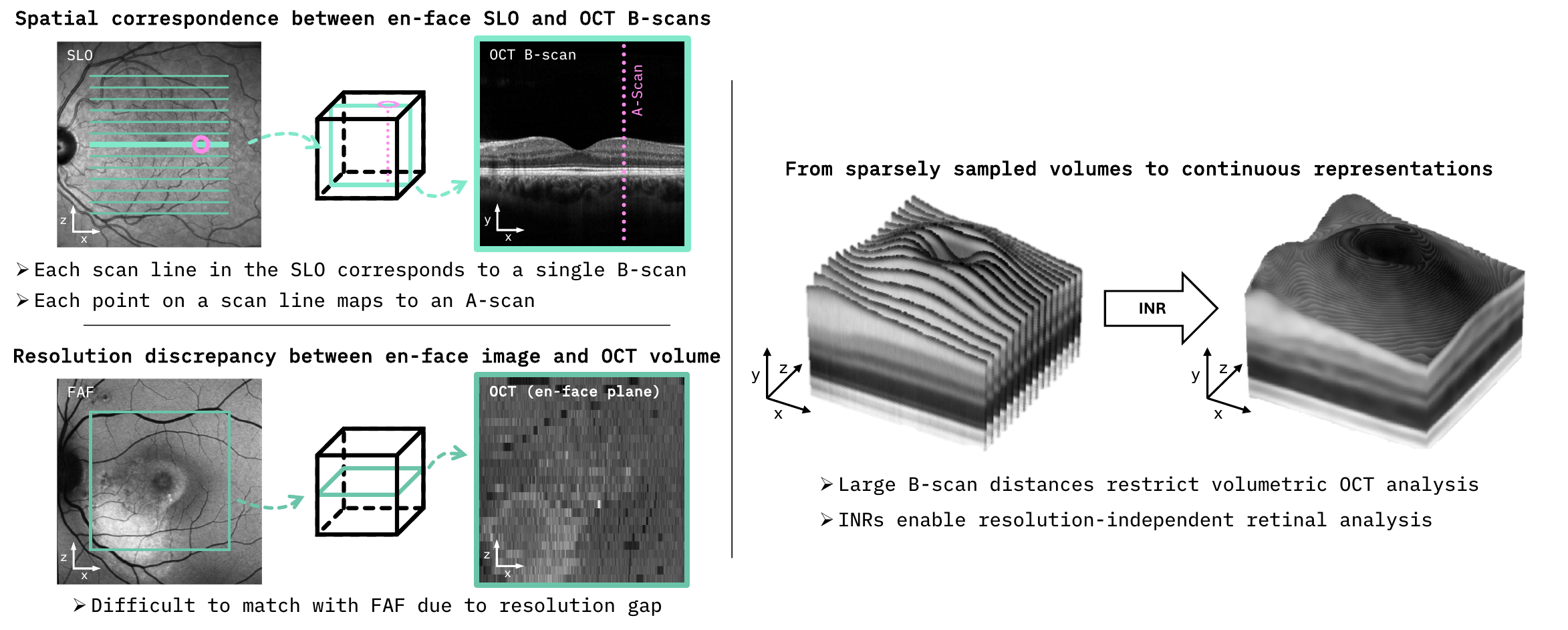}
\caption{Top left: Interrelation between en-face image (here: SLO) and OCT. The position marked in the SLO (\inlinering{barbiepink}{0.6ex}) corresponds to an A-scan (dotted line) in the OCT B-scan. Bottom left: Difference in resolution between an en-face image (here: FAF) and an analogous lateral slice of OCT. The limited amount of B-scans acquired in clinical practice leads to high anisotropy. Right: By leveraging a generalizable INR, we create resolution-independent representations of singular data instances or an atlas, respectively.}
\label{fig:overview}
\end{figure*}

\enluminure{O}{ptical} coherence tomography (OCT) is a standard imaging modality in ophthalmology~\citep{huang1991optical}. Due to its noninvasive nature, it is routinely used to measure alteration in layer thickness and visualize fluid accumulations or other structural changes of the central retina. With this, OCT supports the identification of a broad spectrum of ophthalmological diseases, such as age-related macular degeneration (AMD)~\citep{schmidt2017view}, central serous chorioretinopathy (CSCR)~\citep{chhablani2020multimodal} and diabetic retinopathy~\citep{virgili2015optical}, as well as neurodegenerative conditions like multiple sclerosis \citep{britze2018optical}.   
Systematic learning-based analysis of OCT, whether of single scans, scan series, or population-based statistics, e.g., based on atlases~\citep{chakravarty2018construction}, can enhance diagnostics and therapy in clinical practice. However, the nature of its acquisition brings along some specific challenges:
In favor of shorter acquisition times, clinical routine OCT uses large inter-B-scan distances, leading to sparse sampling in the slow scanning direction and, consequently, to an anisotropic spatial resolution. Most learning-based approaches using convolutional neural networks (CNNs) therefore opt for two-dimensional (2D) solutions rather than performing full three-dimensional (3D) analyses, e.g., for the segmentation of retinal layers~\citep{he2019twoDimOctSeg,pekala2019twoDimOctSeg,li2020deepretina,kepp2023shape}. However, 2D segmentations do not take into account the information from neighboring B-scans, which may lead to inconsistencies between segmentations of individual slices. The resulting 3D surfaces typically are irregular and require extensive post-processing using graph-cut approaches~\citep{li2006optimal}, additional refinement networks~\citep{liu2024simultaneous} or advanced regularization techniques~\citep{Kepp2019_TopologyPreservingShape}. 
Another problem that arises from the sparsely scanned retina is that small anatomical or pathological structures might be missed, reducing the diagnostic precision of OCT image analysis solutions~\citep{velaga2017impact,burchard2024analysis}. Volumetric measurements performed on these anisotropic scans are inherently imprecise since structures appear stretched or compressed. Some structural context for the space between the B-scans is maintained by en-face scanning laser ophthalmoscopy (SLO), which offers high-resolution imaging of the fundus and is used as a reference for eye tracking in certain OCT systems. However, SLO does not contain differentiated depth information. Similarly, other 2D modalities, such as color fundus photography or fundus autofluorescence (FAF), can also hold valuable complementary information, especially about pathological structures visible on the retinal surface. The relation between these modalities and visible biomarkers is yet to be fully explored~\citep{santarossa2022chronological,velaga2022correlation}. The spatial correspondence and the resolution discrepancy between en-face SLO/FAF and OCT B-scans are illustrated in Fig.~\ref{fig:overview} (left, top and bottom).

The resolution-agnostic nature and structural flexibility of implicit neural representations (INRs) suggest that they have the potential to work well in the presence of these challenges.
INRs aim to approximate individual data instances as continuous functions and have gained much traction in research recently. They have been shown to be able to efficiently represent and reconstruct natural images and scenes~\citep{sitzmann2020implicit,mildenhall2021nerf}. With additional instance-specific latent codes or layer modifications, they become generalizable and can be trained population-based to represent multiple images at once~\citep{chen2019learning,park2019deepsdf,kazerouni2024incode,dupont2022data,kim2023generalizable}. Consequently, a lot of research has been done to examine their applicability to medical tasks such as semantic segmentation~\citep{stolt2023nisf}, deformable registration~\citep{wolterink2022implicit} or atlas generation~\citep{grossbrohmer2024sina, dannecker2024cina}. 
However, these medical applications focus on data with far less anisotropy, such as CT or MRI, while highly anisotropic data like OCT remain rather underexplored. 

\paragraph{Our contribution} In this paper, we explore the usage of INRs for OCT-specific tasks focused on the challenging anisotropy of the data (Fig.~\ref{fig:overview}, right). 1) We build upon previous work~\citep{kepp2025bridging} and make use of densely sampled but two-dimensional SLO and FAF information for the interpolation in-between B-scans. This increases the resolution in the slow scanning direction and is applied to images from healthy and diseased subjects and called anisotropic interpolation hereafter. We leverage a generalizable INR to incorporate the additional en-face modality and support this task through population-based retinal layer or pathology segmentation. 2) We perform implicit inter-subject registration and combine it with the introduced generalization approach to perform population-based registration and create an intensity- and shape-based atlas. An additional INR network enables the construction of a resolution-independent atlas, making it especially suited for anisotropic image data like OCT.

\section{Related Works}

\subsection{Interpolation of Anisotropic OCT Images} Various interpolation methods exist to enhance anisotropic OCT image data resolution. Simple intensity-based methods are unable to capture shape and localization differences between B-scans, leading to artifacts and ambiguous results. \cite{lindberg_2018_octInterpolation} propose a weighted combination of linear and transfinite interpolation to achieve an improved interpolation of OCT images, but rely on two OCT images taken with orthogonal scan directions. Registration-based interpolation methods~\citep{ehrhardt2007structure} align structural features more accurately and reduce misalignment artifacts, but cannot infer information missing between slices, limiting reconstruction quality in sparsely sampled regions. Generative approaches for super-resolution are capable of producing realistic, detailed images by reconstructing fine textures and complex structures, e.g. \citep{lopez2023SuperResGAN}, and thus constitute an approach to B-scan interpolation as well. However, most GAN-based approaches for OCT super-resolution only perform B-scan super-resolution but do not interpolate between them~\citep{huang_2019_octSuperRes,das_2020_octSuperRes,yuan_2023_octSuperRes}, or require pairs of high- and low-resolution images~\citep{huang_2019_octSuperRes,yuan_2023_octSuperRes}. Furthermore, generative approaches require a large amount of training data and carry the risk of hallucinations, making them precarious in clinical context~\citep{cohen2018distribution}. 

A limited number of previous works have already used INRs for optical imaging.
\cite{li2025computational} built upon the capability for multi-view scene reconstruction demonstrated by neural radiance fields (NeRF) \citep{mildenhall2021nerf} and proposed a ray tracing-based framework for multi-view reconstruction for OCT of different animal anatomies. Similarly, \cite{xiao2025limited} used an INR for limited view reconstruction of photoacoustic scans of vessels in mice. These works showed that INRs have the capability to work well with sparsely acquired optical imaging data, and the advances of NeRF can be applied well to the medical field. However, methods for multi-view reconstruction cannot directly be applied to interpolation, since slices have no varied angles or overlaps.

\subsection{INR Super-Resolution}
While CNN-based approaches are limited to the resolution of the training data, INRs are resolution-independent, as they deliver continuous representations of the imaged objects. Therefore, INRs inherently are a well-suited choice for the task of super-resolution -- which is methodologically close to the here performed B-scan interpolation. Super-resolution of 3D medical images using INRs can, for example, be found in~\citep{mcGinnis2023inrSuperRes,wu2022arbitrary,fang2024cycleinr}. \cite{wu2022arbitrary} use a CNN encoder and an INR decoder to generate high-resolution MR images from the respective low-resolution images. Similarly, Fang. et al. use a CNN-based encoder to extract feature vectors from low-resolution CT images. These vectors are fed to an INR together with the 3D coordinates of the high-resolution image to predict the pixel intensities for the higher resolution~\citep{fang2024cycleinr}. \cite{mcGinnis2023inrSuperRes} make use of multi-contrast MRI, in which different low-resolution contrasts are acquired in different directions, to produce densely sampled, high-resolution images for single subjects. 

While these methods either work with low-resolution but equidistantly sampled images or use more than one anisotropic image of the same scene, the resolution-agnostic nature of INRs also directly allows processing single an\-i\-so\-tro\-pic images. However, with very large inter-slice distances, as frequently used in retinal OCT imaging, fitting an INR to a single OCT volume results in overfitting to the positions given in the training data. That is, meaningful results are generated for the B-scan positions used during training, but there is no useful interpolation between the B-scans. In~\citep{amiranashvili2022learning}, this overfitting problem is overcome with a generalizable INR, which is trained on highly an\-i\-so\-tro\-pic segmentations. An MLP is trained that is shared between all training instances, while case-specific latent priors are adapted for each sample, allowing the generation of dense and smooth shapes. Building on this approach, \cite{stolt2023nisf} train a generalizable INR to reconstruct and segment short-axis cardiac MRI and show smooth segmentations on long-axis images. 

\subsection{OCT Atlases}
Equivalent to less anisotropic modalities, an OCT atlas can be constructed by combining deformable registration and optimization of an atlas image. In the context of optical imaging, an OCT atlas can be helpful to differentiate between physiological variability in normative tissue and pathological alterations. Based on these differences, many diseases have been shown to be classifiable using atlases.
\cite{chakravarty2018construction} created a normative atlas and calculated deviation to classify AMD. They proposed to use a geometric mean image as an initial template and employ discrete registration for computational efficiency due to the large size of OCT B-scans. \cite{lee2017atlas} employed geometry-based registration to create an atlas and visualized thinning of retinal nerve fiber layers due to glaucoma. \cite{khansari2019automated} create an atlas based on pathological volumes to measure tissue contraction and expansion due to diabetic retinopathy.

In general, these previous approaches demonstrated the value of correctly aligning individual layers during the construction of the atlas, sometimes through explicit conditions on layer shapes.
However, each of these approaches is limited by the resolution of the original OCT volumes. Consequently, the atlas is highly restrictive regarding the data it is applicable to. With different acquisition setups in clinical practice, this is an important factor to overcome.
In this work, we will focus on the methodology to make the atlas generation resolution-agnostic and broadly applicable across varying data settings.

\subsection{INR Registration and Atlases}
\cite{wolterink2022implicit} introduced implicit deformable image registration, where a deformation field is represented by an INR and optimized for a single image pair. There have since been multiple adaptations proposed, e.g., B-spline regularization~\citep{sideri2024sinr} or inverse consistency~\citep{van2023robust,tian2024nephi}. 
To the best of our knowledge, the applicability of \cite{wolterink2022implicit} to OCT or similarly anisotropic data has not been explored in the academic literature yet.

INRs can further be used to represent an image atlas through different approaches. \cite{grossbrohmer2024sina} combine conventional registration with a single INR to represent the atlas image, whereas \cite{dannecker2024cina} use a generalization approach to represent multiple images with one INR and thus omit any registration. It is still an open question how well the generalization capability of population-trained INRs translates to medical optical imaging with the associated challenges of sparsely sampled data, small datasets, and low signal-to-noise ratio.


\section{Materials and Methods}

\begin{figure*}[t]
\includegraphics[width=\textwidth]{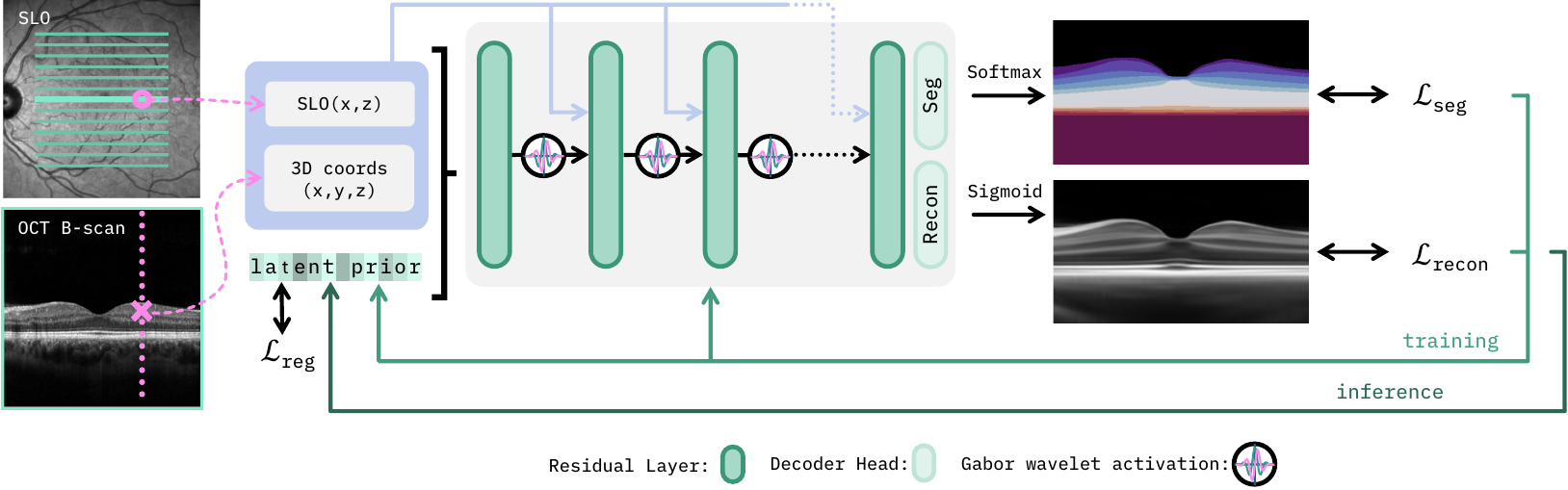}
\caption{Schematic overview of the proposed interpolation framework. The input for the INR are 3D coordinates in the OCT (\inlinecross{barbiepink}{0.6ex}{1.5ex}) and intensities at the corresponding SLO position~(\inlinering{barbiepink}{0.636ex}). The INR is trained to reconstruct the intensity and segmentation label of the OCT volume at the respective coordinate. The reconstructed OCT can be evaluated at any arbitrary coordinate with a known SLO intensity. To reconstruct and segment an unseen case, the INR is kept frozen, and only the latent prior is adapted. }
\label{fig:interp_framework}
\end{figure*}

\subsection{Datasets}
We use an in-house dataset consisting of paired OCT volumes and SLO images from the left and right eyes of 50 healthy volunteers (one OCT volume and one SLO image per eye), resulting in a total of 100 image pairs, to develop, train, and evaluate generalizable INRs for OCT interpolation, segmentation, and atlas generation. All OCT volumes were acquired with a Spectralis OCT scanner (Heidelberg Engineering) modified for research purposes to allow arbitrary inter-B-scan spacing, but image acquisition was performed using the angiography acquisition (OCT-A) mode consistent with standard clinical OCT systems. Acquisition in the angiography mode results in slightly reduced B-scan image quality, as fewer repeated scans per position are averaged compared to structural OCT acquired with typical clinical protocols, but provides densely sampled OCT volumes of size $496\times512\times512$ voxels covering a field of view of $2\times6\times6$\;mm$^3$. In Spectralis scanners, an SLO (scanning laser ophthalmoscopy) is used for eye-tracking and acquired routinely alongside the OCT. The SLO images have a size of $768\times768$ pixels, covering a larger field of view in the lateral plane (approx. $8.6\times8.6$\;mm$^2$) than the OCT images. To match OCT and SLO, we use the localization information provided by the OCT device to subsample the SLO image at the A-scan positions corresponding to each B-scan. In the OCT images, twelve layers (eleven retinal layers plus choroid) were segmented automatically using the Iowa Reference Algorithms~\citep{li2006optimal}. Subsequently, segmentation errors were corrected manually by an expert. For preprocessing, all B-scans are flattened in relation to the Bruch's membrane and cropped to $230\times512$~pixels so that the retina is at the center height of the image. Images of right eyes are flipped for identical orientation of all images.

Additional experiments are performed on a clinical dataset containing images from 19 patients with central serous chorioretinopathy (CSCR). The CSCR dataset contains 146 OCT volumes consisting of 25 B-scans with a resolution of $496\times512$ pixels. Like the healthy images, the CSCR images span a field of view of $2\times6\times6$\;mm$^3$. As an additional en-face modality, FAF (fundus autofluorescence) images are used for the CSCR data. They have the same size ($496\times512\times512$~voxels) and field of view ($8.6\times8.6$\;mm$^2$) as the SLO images. To match the FAF images to the OCT volumes, we use a two-step procedure utilizing the accompanying SLO. First, the FAF image is registered onto the SLO using the MedRegNet registration pipeline~\citep{santarossa2022medregnet}, with improved RANSAC as proposed in~\citep{santarossa2025robust}. Next, the registered FAF is subsampled to the OCT resolution using the same SLO localization information as before. This entire two-step procedure, albeit in reverse (i.e., mapping OCT onto FAF), has been detailed in~\citep{santarossa2022chronological}.
For the CSCR data, the retina as well as pathologies (intra- and subretinal fluid and pigment epithelial detachment) were segmented manually by experienced ophthalmologists.


\subsection{Anisotropic Interpolation}
To reduce the difficulties arising from the anisotropy of clinical OCT volumes, we build upon the framework for joint inter-B-scan interpolation and segmentation of retinal OCT first proposed in \citep{kepp2025bridging}. The method is shown exemplarily for healthy images in Fig.~\ref{fig:interp_framework}. A generalizable INR, extending the methods proposed by \cite{amiranashvili2022learning} and \cite{stolt2023nisf}, is used that consists of a multi-layer perceptron (MLP) shared between all training samples and learnable subject-specific priors that condition the MLP on the respective subject. Since small structures between B-scans cannot be represented by the sparsely scanned OCT images alone, we propose to use additional en-face imaging modalities as input to the INR. Here, we exemplarily use SLO and FAF images that are densely scanned in the en-face plane. They therefore provide more detailed information about the localization of structures in the lateral direction than OCT. Although these images have no axial depth, we hypothesize that the lateral information enables the INR to represent structures that are not captured in the OCT scanning protocol and are therefore not visible in the B-scans. The INR $f_{\theta}=(f_{\theta}^{\mathtt{recon}},f_{\theta}^{\mathtt{seg}})$ with trainable parameters $\theta$ has two separate output layers, one for the reconstruction of the OCT volume and one for the segmentation of anatomical or pathological retinal structures, and is trained with OCT and corresponding en-face images from $n$ subjects. The network receives three inputs: a 3D coordinate $(x,y,z)\in\mathbb{R}^3$ corresponding to a voxel position in the OCT volume $\mathtt{OCT}_i$ of the subject $i$, the intensity of the en-face image $\mathtt{EF}_{i}(x,z)$ at the respective lateral position $(x,z)$, and the subject-specific prior $\boldsymbol{p}_{i}\in\mathbb{R}^{L}$. 

For one forward pass, we use the coordinates $(x,y,z)$ of an entire OCT B-scan and compute the loss
\begin{align}
    \nonumber \mathcal{L}=&\sum_{i=1}^{n}\sum_{(x,y,z)} [\mathcal{L}_{\mathtt{recon}}(f_{\theta}^{\mathtt{recon}}(x,y,z,\mathtt{EF}_{i}(x,z),\boldsymbol{p}_{i}), \\ \nonumber &\mathtt{OCT}_{i}(x,y,z))+\alpha\mathcal{L}_{\mathtt{seg}}(f_{\theta}^{\mathtt{seg}}(x,y,z,\mathtt{EF}_{i}(x,z),\boldsymbol{p}_{i}), \\&\mathtt{SEG}_{i}(x,y,z))+\beta\mathcal{L}_{\mathtt{reg}}(\boldsymbol{p}_{i})],
\end{align}
to adapt the network parameters and the latent priors simultaneously. Here, $\mathtt{SEG}_{i}(x,y,z)$ is the one-hot encoded ground truth segmentation label at position $(x,y,z)$. A combination of MSE and SSIM loss is used for the reconstruction loss $\mathcal{L}_{\mathtt{recon}}=0.1\cdot\mathcal{L}_{\text{SSIM}} + \mathcal{L}_{\text{MSE}}$ and the binary cross-entropy loss is used for the segmentation loss $\mathcal{L}_{\mathtt{seg}}$. As proposed by \cite{stolt2023nisf}, an L2 regularization loss $\mathcal{L}_{\mathtt{reg}}$ is applied to penalize large values in the latent codes. The loss components are weighted by $\alpha=\beta=0.2$.

After completion of the population-based training, the INR can be fit to an unseen subject $n+1$ by freezing the network $f_\theta$ and adapting a new latent prior $\boldsymbol{p}_{n+1}$. Since for new subjects, typically the OCT image $\mathtt{OCT}_{n+1}(x,y,z)$ is given but not the corresponding segmentation, the loss for learning the new latent code is reduced to 
\begin{align}
    \mathcal{L}=\sum_{(x,y,z)} &\mathcal{L}_{\mathtt{recon}}(f_{\theta}^{\mathtt{recon}}(x,y,z,\mathtt{EF}_{n+1}(x,z),\boldsymbol{p}_{n+1}), \nonumber \\
    &\mathtt{OCT}_{n+1}(x,y,z))+\beta\mathcal{L}_{\mathtt{reg}}(\boldsymbol{p}_{n+1}).
\end{align}
As shown in our previous work~\citep{kepp2025bridging}, learning the latent priors for unseen cases this way results in plausible segmentation predictions. The learned latent prior $\boldsymbol{p}_{n+1}$ enables both the interpolation of OCT intensities and segmentation labels for any coordinate $(x,y,z)$ for which the intensity of the additional 2D modality $\mathtt{EF}_{n+1}(x,z)$ is known. To do so, a simple forward pass is used to calculate $f_{\theta}(x,y,z,\mathtt{EF}_{n+1}(x,z),\boldsymbol{p}_{n+1})$.

\begin{figure*}[t]
    \centering
    \includegraphics[width=\linewidth]{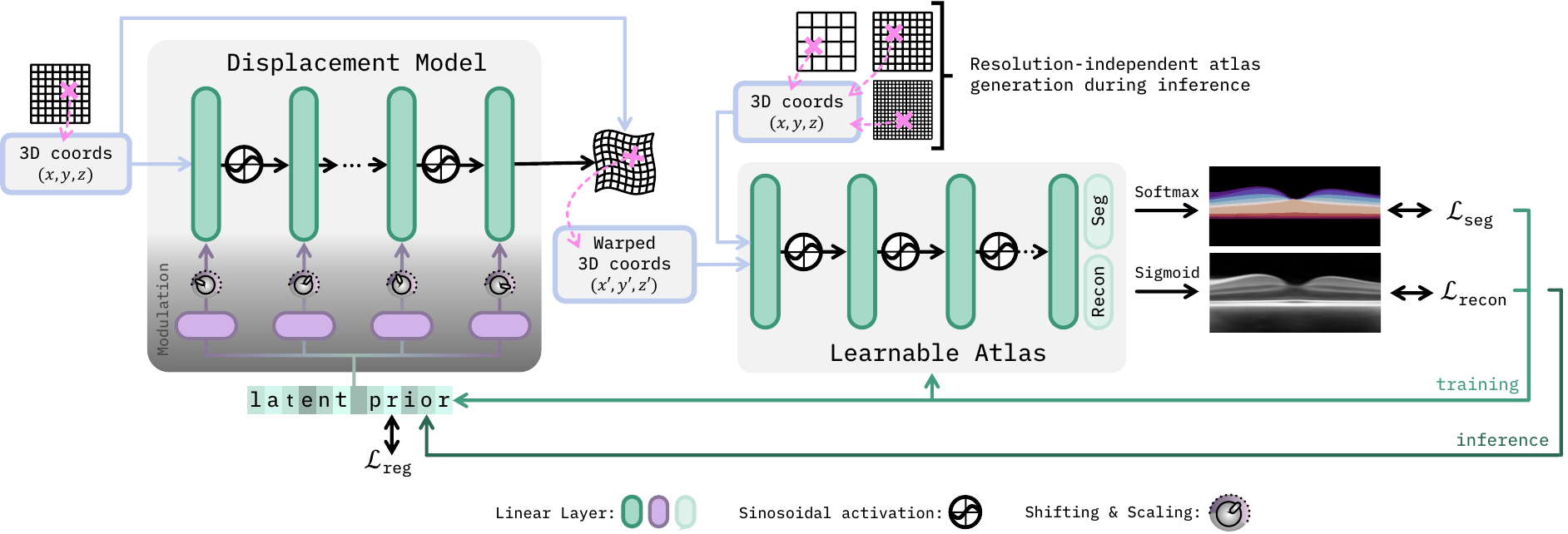}
    \caption{Schematic overview of the proposed atlas generation framework. The generalizable INR predicts a deformation field based on 3D coordinates (\inlinecross{barbiepink}{0.6ex}{1.5ex}) and an instance-specific latent prior. The INR layers are modulated through scale and shift vectors predicted from the latent prior by a hypernetwork (purple layers).  The deformed coordinates are then input into another INR that represents the atlas with intensities and segmentation labels. The learned atlas is resolution-independent and can be evaluated on any arbitrary number of coordinates.}
    \label{fig:reg_atlas_framework}
\end{figure*}

\subsubsection{Implementation Details}
The generalizable INR for OCT interpolation consists of a six-layer MLP using complex Garbor wavelet activation functions~\citep{saragadam2023wire} plus two separate linear layers as decoder heads for reconstruction output $f_{\theta}^{\mathtt{recon}}$ and segmentation output $f_{\theta}^{\mathtt{seg}}$. All hidden layers have 128 nodes, and the reconstruction head outputs scalar grayscale values, while the segmentation head generates $C$ output values for one-hot encoded class labels, where $C$ is the number of classes considered in the respective task. That is, for the segmentation of retinal layers, twelve labels corresponding to the different tissue layers -- including the choroid -- are predicted, and for the segmentation of pathological images, three labels corresponding to the background, retina, and pathologies are predicted. The reconstruction output layer uses sigmoid activation, whereas the segmentation head uses softmax activation. All hidden layers use residual connections, and to enhance focus on input coordinates as well as the additional imaging modality, we pass these inputs repeatedly to the hidden layers. The length~$L$ of the latent vectors is set to 128 empirically. Network training is performed for 1,500 epochs, where one epoch corresponds to a run over all B-scans of the training images. We use an exponential learning rate decay of 0.99 and an initial learning rate of \textexp{10}{-4}. Early stopping is enabled after 100 epochs using a patience of 20 epochs and a minimum loss improvement of \textexp{10}{-5}. The generalizable INR was trained for up to three days on a single NVIDIA GeForce GTX 1080 Ti.

\subsection{Registration and Atlas Generation}

To overcome the resolution dependence of retinal atlases, we continue to use an INR for efficient handling of the anisotropic data. We combine INR-based inter-subject registration with atlas generation -- enabling joint optimization of registration and atlas without meta-learning. Our proposed method consists of two key components that jointly model the deformation from the atlas to the subject space, as well as the atlas representation itself. These components are 1) a learnable atlas volume encoded with an INR and 2) a generalizable INR that models the deformation fields with latent priors that modulate the generalizable INR to capture subject-specific variations. The joint framework is shown in Fig.~\ref{fig:reg_atlas_framework} and is described in detail below. The resulting framework takes 3D coordinates and a latent prior as input to predict the deformed atlas values with auxiliary information, in our case retina layer labels.

\textbf{The learnable atlas} is a key component to enable resolution independence. It builds upon the idea proposed by \cite{grossbrohmer2024sina} of representing the atlas using an MLP $f_\theta$ with adjustable parameters $\theta$. As such, the atlas is trained using warped input coordinates to predict the most similar subject representation. In this work, we extend this approach by incorporating a combined representation of intensity values and segmentation labels, similar to our previously proposed interpolation method. Thus, we introduce an INR $f_{\theta}=(f_{\theta}^{\mathtt{recon}},f_{\theta}^{\mathtt{seg}})$, which takes warped 3D coordinates $(x',y', z')\in\mathbb{R}^3$ to reconstruct the intensity value of the OCT image and the class label of the retinal layers. Since we aim to construct a single atlas volume that represents the whole dataset, this INR is (different from our previous approach) non-generalizable. This continuous representation of the atlas enables the intensity and segmentation values to be sampled instantly, without the need for explicit interpolation in the registration process. 

\textbf{The displacement model} is a generalizable INR, denoted as $g_\vartheta$ with trainable parameters $\vartheta$. In contrast to the proposed method for interpolation, the displacement model uses a hypernetwork to modulate its interim outputs, similar to \cite{dannecker2024cina} and \cite{dupont2022data}. To do so, we use the hypernetwork (depicted in purple in Fig.~\ref{fig:reg_atlas_framework}) -- consisting of one linear layer for each layer of the INR (depicted in turquoise in Fig.~\ref{fig:reg_atlas_framework})  -- to predict individual scaling and shifting vectors based on the latent prior input $\boldsymbol{p}_{i} \in \mathbb{R}^L$. Let $H$ be the hidden size of the INR. The individual scaling vector ~$\boldsymbol{\phi}\in \mathbb{R}^H$ and shifting vector~$\boldsymbol{\psi}\in \mathbb{R}^H$ are the result of the linear mapping $(\boldsymbol{\phi},\boldsymbol{\psi})^T =W_{\text{hyper}} \boldsymbol{p}_i + \boldsymbol{b}_{\text{hyper}}$ with weight matrix~$W_{\text{hyper}}\in \mathbb{R}^{2H\times L}$, and bias~$\boldsymbol{b_{\text{hyper}}}\in \mathbb{R}^{2H}$.
By modulating $g_\vartheta$ with scaling vector~$\boldsymbol{\phi}$ and shifting vector~$\boldsymbol{\psi}$, the output of a linear layer $\sigma (W \boldsymbol{x} + \boldsymbol{b})$ with input~$\boldsymbol{x}$, weight matrix $W$, bias $\boldsymbol{b}$ and activation function $\sigma$ becomes $\sigma(\boldsymbol{\phi} (W\boldsymbol{x}+\boldsymbol{b}) + \boldsymbol{\psi})$.

The resulting latent code-driven architecture is then used to represent a continuous displacement field $u_{i}: \mathbb{R}^3 \rightarrow \mathbb{R}^3$ for each subject $i$ as already introduced for classic INRs by \cite{wolterink2022implicit}.
Thus, this model  gets the original 3D input coordinates $(x,y,z)\in\mathbb{R}^3$ and a latent prior $\boldsymbol{p}_{i}\in\mathbb{R}^{L}$. For each coordinate $(x,y,z)$, we thereby obtain a subject-specific displacement 
\begin{align}
u_{i}(x,y,z) = g_{\vartheta}(x,y,z,\boldsymbol{p}_{i})
    \label{eq:u}
\end{align}
and warped coordinates
\begin{align}
    \varphi_{i}(x,y,z) = u_{i}(x,y,z) + (x,y,z)
    \label{eq:varphi}
\end{align}
that act as the input $(x',y',z')$ to the previously described learnable atlas.

In contrast to the architecture of \cite{dannecker2024cina}, we separate the parts $\boldsymbol{p}_{i_1}$ and $\boldsymbol{p}_{i_2}$ of the latent prior $\boldsymbol{p}_{i} = (\boldsymbol{p}_{i_1}, \boldsymbol{p}_{i_2})$ that predict $\boldsymbol{\phi}$ and $\boldsymbol{\psi}$ and use two normal distributions $\boldsymbol{p}_{i_1} \sim \mathcal{N}(1,0.1) $ and $\boldsymbol{p}_{i_2} \sim \mathcal{N}(0,0.1)$ to initialize them. With $\boldsymbol{\phi} = 1$ and $\boldsymbol{\psi} =0$ the resulting generalizable INR corresponds to a classic INR. This initialization therefore reduces the disturbance from a non-trained latent prior right after initialization.

Training and inference for atlas generation work analogously to those we used for B-scan interpolation. Assembled into one framework, a subset of coordinates $(x,y,z)$ of an OCT volume is used to compute the reconstruction and alignment error as
\begin{align}
    \mathcal{L}=&\sum_{i=1}^{n}\sum_{(x,y,z)} [\mathcal{L}_{\mathtt{recon}}(f_{\theta}^{\mathtt{recon}}(\varphi_{i}(x,y,z)), \mathtt{OCT}_{i}(x,y,z)) \nonumber \\ 
    &+\alpha\mathcal{L}_{\mathtt{seg}}(f_{\theta}^{\mathtt{seg}}(\varphi_{i}(x,y,z)), \mathtt{SEG}_{i}(x,y,z)) \nonumber \\ 
    &+\beta\mathcal{L}_{\mathtt{reg}}(u_{i}(x,y,z))].
\end{align}
Note that this loss is dependent on the network parameters $\vartheta$ and the latent priors $\boldsymbol{p}_1,\dots,\boldsymbol{p}_n$ through Eq.~\eqref{eq:u} and Eq.~\eqref{eq:varphi}.
$\mathtt{OCT}_{i}(x,y,z)$ and $\mathtt{SEG}_{i}(x,y,z)$ are the ground truth intensity or segmentation labels at position $(x,y,z)$. The reconstruction loss $\mathcal{L}_{\mathtt{recon}}$ is based on the image intensity and uses MSE loss. Binary cross-entropy loss is used as the segmentation loss $\mathcal{L}_{\mathtt{seg}}$. To regularize the deformation, L1 loss is used as $\mathcal{L}_{\mathtt{reg}}$. The weighting factors are experimentally chosen and set to $\alpha = 1.0$ and $\beta = 0.01$.

To apply the model to previously unseen subjects, only the latent prior $\boldsymbol{p}_{n+1}$ needs to be optimized, while the deformation model $g_\vartheta$ and the atlas representation $f_\theta$ stay frozen. Since the available information during inference is limited to the OCT intensity values of one subject, the loss function is modified to 
\begin{align}
    \mathcal{L}=&\sum_{(x,y,z)} [\mathcal{L}_{\mathtt{recon}}(f_{\theta}^{\mathtt{recon}}(\varphi_{i}(x,y,z)), \mathtt{OCT}_{n+1}(x,y,z)) \nonumber \\&+\gamma\mathcal{L}_{\mathtt{reg}}(u_{i}(x,y,z))]
\end{align}
where $\gamma$ is a weighting factor and set to $0.01$.

\begin{table*}[htbp]
    \centering
    \caption{OCT interpolation and segmentation results for the healthy OCT data. Results are reported for our proposed generalizable INR with (GenINR$_\text{SLO}$) and without (GenINR) additional SLO input for train and test data separately. For comparison, interpolation results are reported for linear and registration-based interpolation, as well as for standard INRs adapted to single images with and without SLO guidance (SingleINR$_\text{SLO}$ / SingleINR), evaluated on the training data only since these instance-based methods require ground truth annotations for some of the B-scans. Metrics are stated as the \meanstd{mean}{std}. The best results are highlighted. All INR methods show significant differences in Dice, ASSD, and SSIM in comparison to the registration-based interpolation ($p < 0.001$, Wilcoxon signed-rank test with Holm-Bonferroni correction).}
    \begin{tabular}{l@{\extracolsep\fill}llllllll}
        \hline
         \hspace{6mm} & & \multicolumn{4}{l}{Image Reconstruction} & \multicolumn{3}{l}{Image Segmentation} \\
         & Method & $\downarrow$MAE$^{[\%]}$ & $\uparrow$PSNR &  $\uparrow$SSIM$^{[\%]}$ & $\downarrow$LPIPS & $\uparrow$Dice$^{[\%]}$ & $\downarrow$ASSD$^{[\SI{}{\micro\metre}]}$ &  $\downarrow$HD$^{[\SI{}{\micro\metre}]}$ \\ \hline
         \parbox[t]{3mm}{\multirow{6}{*}{\rotatebox[origin=c]{90}{train}}} 
         & linear & \meanstd{5.92}{0.55} & \meanstd{21.7}{0.5} & \meanstd{39.7}{5.6} & \meanstd{$\textbf{0.12}$}{0.03} & \meanstd{88.8}{1.1} & \meanstd{9.3}{0.7} & \meanstd{37.2}{3.3}\\
         & reg. & \meanstd{5.76}{0.55} & \meanstd{22.0}{0.5} & \meanstd{41.3}{5.6} & \meanstd{$\textbf{0.13}$}{0.03} & \meanstd{89.1}{1.2} & \meanstd{7.5}{0.8} & \meanstd{39.2}{4.4} \\
         & SingleINR & \meanstd{6.65}{0.64} & \meanstd{20.6}{0.5} & \meanstd{36.7}{5.9} & \meanstd{0.26}{0.02} & \meanstd{90.8}{1.1} & \meanstd{7.7}{0.7} & \meanstd{37.6}{3.6}\\
         & SingleINR$_\text{SLO}$ & \meanstd{9.42}{1.51} & \meanstd{17.5}{1.0} & \meanstd{26.8}{6.4} & \meanstd{0.45}{0.07} & \meanstd{83.6}{5.3} & \meanstd{18.7}{12.0} & \meanstd{603.4}{179.8} \\
         & GenINR & \textbf{\meanstd{5.24}{0.51}} & \textbf{\meanstd{22.6}{0.5}} & \textbf{\meanstd{48.9}{5.7}} & \meanstd{0.53}{0.03} & \textbf{\meanstd{91.9}{1.0}} & \textbf{\meanstd{7.0}{0.7}} & \textbf{\meanstd{32.5}{3.7}} \\
         & GenINR$_\text{SLO}$ & \textbf{\meanstd{5.24}{0.51}} & \textbf{\meanstd{22.6}{0.5}} & \textbf{\meanstd{48.9}{5.7}} & \meanstd{0.52}{0.03} & \textbf{\meanstd{91.9}{1.0}} & \textbf{\meanstd{7.0}{0.6}} & \textbf{\meanstd{32.5}{3.7}} \\
         &  &  &  &  &  &  &  &  \\
         \parbox[t]{3mm}{\multirow{2}{*}{\rotatebox[origin=c]{90}{test}}} 
         & GenINR & \meanstd{6.04}{0.57} & \meanstd{21.4}{0.5} & \meanstd{44.3}{6.0} & \meanstd{0.55}{0.03} & \meanstd{86.5}{3.7} & \meanstd{11.4}{2.1} & \meanstd{45.1}{4.9} \\
         & GenINR$_\text{SLO}$ & \meanstd{6.08}{0.59} & \meanstd{21.4}{0.5} & \meanstd{44.2}{6.0} & \meanstd{0.54}{0.03} & \meanstd{86.2}{3.8} & \meanstd{11.7}{2.2} & \meanstd{45.4}{5.0} \\ \hline
    \end{tabular}
    \label{tab:interp_results_healthy}
\end{table*}

\subsubsection{Implementation Details}
The atlas representation $f_\theta$ is a four-layer MLP with two additional separate layers for reconstruction $f_{\theta}^{\mathtt{recon}}$ and segmentation $f_{\theta}^{\mathtt{seg}}$. Each hidden layer has 256 nodes and uses a sinusoidal activation function as proposed by \cite{sitzmann2020implicit}. The prediction heads are based on sigmoid activation for the image intensity values and softmax activation to predict the class values as one-hot encoding. Different from the interpolation task, here, the class labels are redefined to ten retinal layers by merging very thin layers and discarding the choroid label. While \cite{grossbrohmer2024sina} suggest using a warm-start procedure by increasing the learning rate, we propose to pre-train the atlas INR $f_\theta$ for 500 epochs to represent the voxel-wise median of the dataset, resulting in a less blurred initialization. The generalizable INR used to model the deformation consists of four layers with 128 nodes and uses sinusoidal activations throughout, including a single sine activation in the final layer to predict the three-dimensional displacement vector. The hypernetwork's layers are three separate linear layers with 256 nodes each. The model weights are initialized by sampling from a uniform distribution $W \sim \mathcal{U}(-0.001, 0.001)$ for small and smooth initial deformations. 
The vector length of 128 for the latent space is chosen empirically. All components -- the deformation model, the atlas model, and the latent priors -- are trained simultaneously for 2,000 iterations using the Adam optimizer with learning rates of \textexp{10}{-5}, \textexp{10}{-6}, and \textexp{10}{-5}, respectively. The optimizer is combined with an exponential learning rate scheduler with a multiplication factor of 0.999. The batch size is set to one subject, with 916,608 coordinates each. This number corresponds to an equidistant downsampling to a sub-volume containing 231$\times$62$\times$64~voxels. This volume is randomly selected from all shifted variations of downsampled sub-volumes. Training was conducted on two NVIDIA RTX A5000s and completed in approximately eight hours. During inference, one latent prior is trained for each subject for 100 epochs with a learning rate of \mbox{5$\times$\textexp{10}{-4}} and the same exponential learning rate scheduler as for training. 

\begin{table*}[htbp]
    \centering
    \caption{OCT interpolation and segmentation results for the CSCR data. We compare the proposed generalizable INR with and without FAF input (GenINR$_\text{FAF}$, GenINR) to linear and registration-based interpolation as well as INRs adapted to single images (SingleINR / GenINR). Results are again reported separately for train and test images. Metrics are stated as the \meanstd{mean}{std}. The best results are highlighted. All INR methods show significant differences in $\text{Dice}_\text{retina}$, $\text{Dice}_\text{fluid}$, and SSIM in comparison to the registration-based interpolation ($p < 0.001$, Wilcoxon signed-rank test with Holm-Bonferroni correction).} 
    \begin{tabular}{l@{\extracolsep\fill}lllllll}
        \hline
         \hspace{6mm} & & \multicolumn{4}{l}{Image Reconstruction} & \multicolumn{2}{l}{Image Segmentation} \\
         & Method & $\downarrow$MAE$^{[\%]}$ & $\uparrow$PSNR &  $\uparrow$SSIM$^{[\%]}$ & $\downarrow$LPIPS & $\uparrow \text{Dice}^{[\%]}_\text{retina}$ & $\uparrow \text{Dice}^{[\%]}_\text{fluid}$  \\ \hline
         \parbox[t]{3mm}{\multirow{6}{*}{\rotatebox[origin=c]{90}{train}}} 
         & linear & \meanstd{6.12}{1.35} & \meanstd{21.30}{1.37} & \meanstd{39.90}{9.06} & \textbf{\meanstd{0.12}{0.03}} & \meanstd{98.05}{0.59} & \meanstd{49.54}{23.15} \\
         & reg. & \meanstd{5.82}{1.28} & \meanstd{21.68}{1.34} & \meanstd{42.55}{9.13} & \meanstd{0.13}{0.03} & \meanstd{98.45}{0.53} & \meanstd{44.04}{24.00} \\
         & SingleINR & \meanstd{6.86}{1.55} & \meanstd{20.23}{1.38} & \meanstd{36.83}{9.80} & \meanstd{0.29}{0.05} & \meanstd{98.49}{0.64} & \meanstd{47.52}{23.05} \\
         & SingleINR$_\text{FAF}$ & \meanstd{7.54}{1.64} & \meanstd{19.32}{1.34} & \meanstd{33.69}{9.51} & \meanstd{0.31}{0.06} & \meanstd{98.08}{0.73} & \meanstd{45.94}{23.39} \\
         & GenINR & \textbf{\meanstd{5.33}{1.24}} & \textbf{\meanstd{22.50}{1.42}} & \textbf{\meanstd{49.16}{9.74}} & \meanstd{0.55}{0.07} & \textbf{\meanstd{98.99}{0.48}} & \meanstd{50.71}{27.92} \\
         & GenINR$_\text{FAF}$ & \textbf{\meanstd{5.34}{1.25}} & \meanstd{22.48}{1.42} & \meanstd{49.11}{9.74} & \meanstd{0.54}{0.07} & \meanstd{98.98}{0.47} & \textbf{\meanstd{51.03}{27.51}} \\
         &  &  &  &  &  &  & \\
         \parbox[t]{3mm}{\multirow{2}{*}{\rotatebox[origin=c]{90}{test}}} 
         & GenINR & \meanstd{6.28}{1.25} & \meanstd{20.25}{1.27} & \meanstd{44.45}{9.41} & \meanstd{0.57}{0.07} & \meanstd{97.53}{1.57} & \meanstd{16.84}{25.86}  \\
         & GenINR$_\text{FAF}$ & \meanstd{6.59}{1.21} & \meanstd{20.29}{1.21} & \meanstd{44.67}{9.50} & \meanstd{0.56}{0.07} & \meanstd{97.56}{1.57} & \meanstd{17.59}{26.75}  \\ \hline
    \end{tabular}
    \label{tab:interp_results_cscr}
\end{table*}

\begin{figure*}[htbp]
        \includegraphics[width=\linewidth]{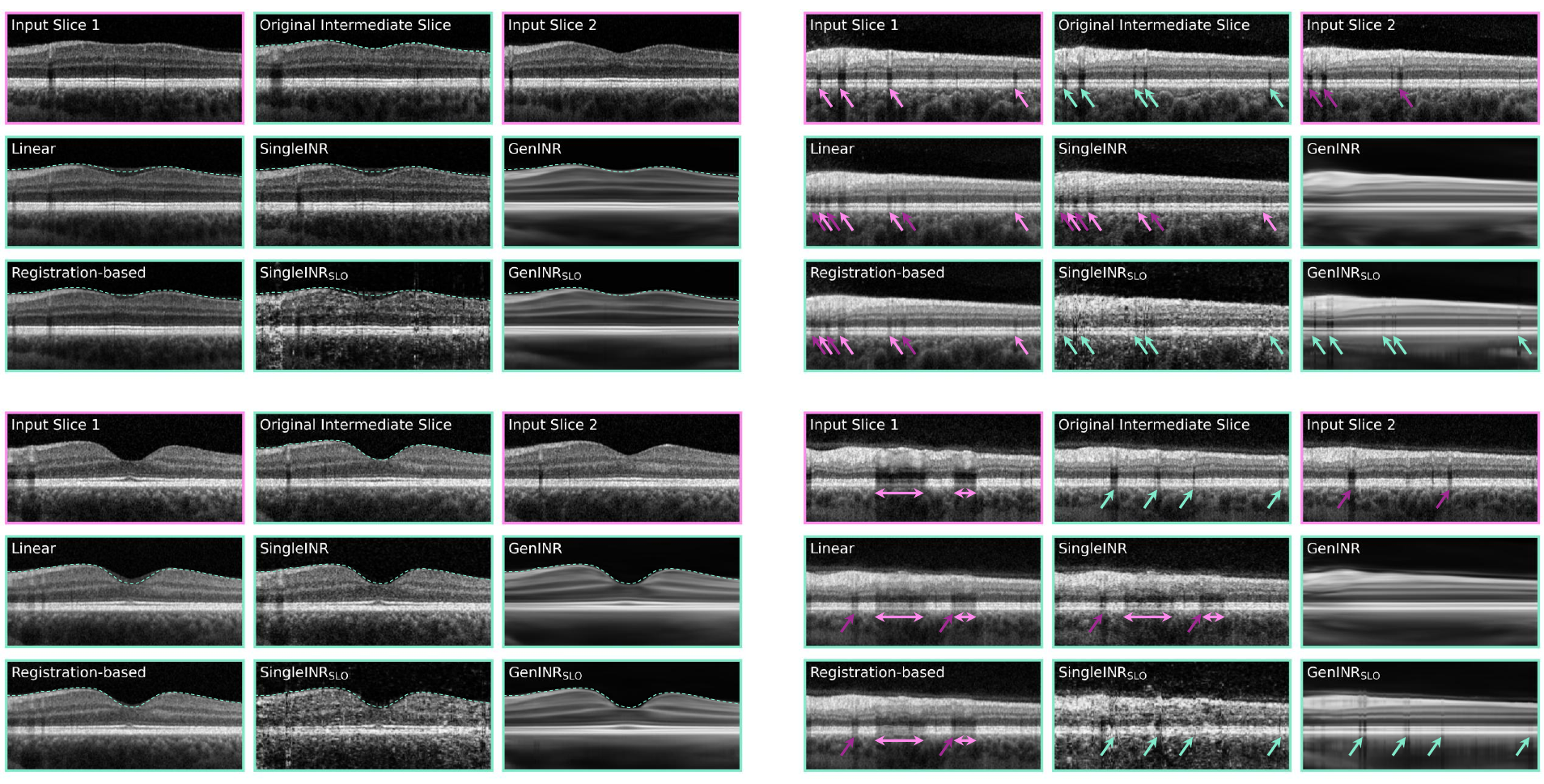}
    \caption{Four examples of interpolated B-scans using linear, registration-based \citep{ehrhardt2007structure}, and INR-based interpolation. The input slices for the different methods are shown with \textcolor{barbiepink}{pink frames}, while the original intermediate and interpolated B-scans are shown with \textcolor{SeaGreen}{turquoise frames}. On the left side, two examples from the foveal region of the retina are depicted, showing interpolation artifacts for instance-based methods, while our GenINR manages to correctly continue the shape of the retina. On the right side, examples from the outer border of the retina are shown. Here, blood vessels are correctly localized by the GenINR with SLO integration (\textcolor{SeaGreen}{turquoise arrows}). For the other methods, vessels can only be propagated from the input slices, again leading to interpolation artifacts.}
    \label{fig:qualitative_results_slo}
\end{figure*}

\begin{figure*}[htbp]
        \includegraphics[width=\linewidth]{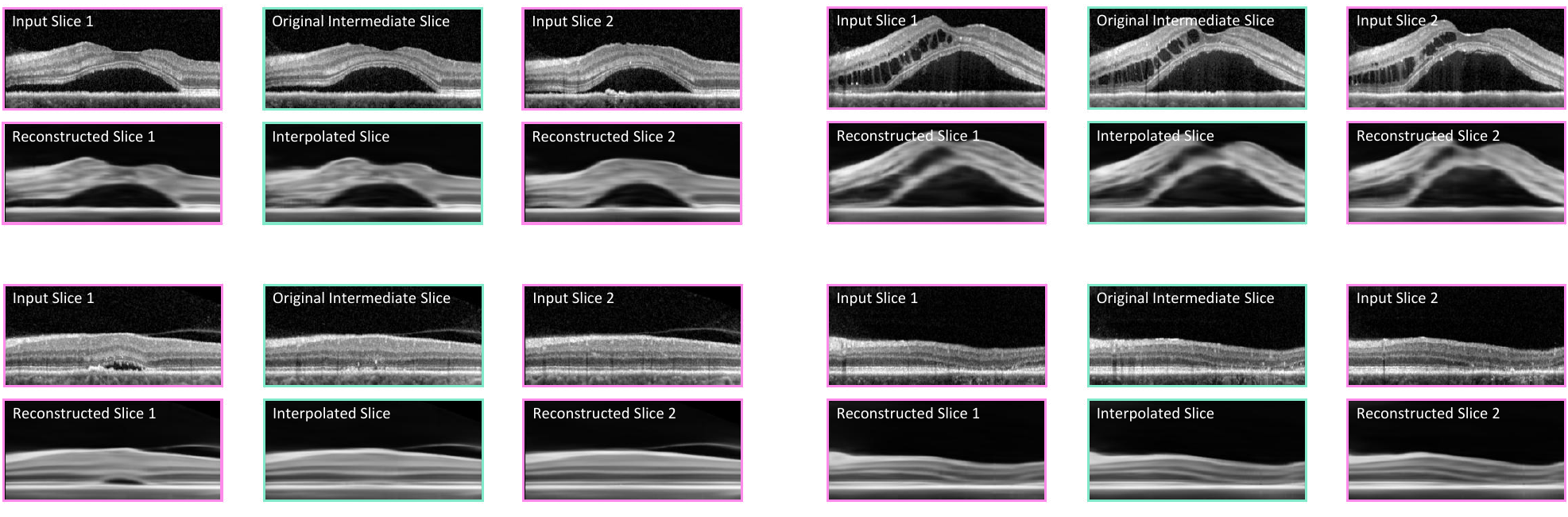}
    \caption{Four examples of interpolated B-scans from a CSCR patient using the proposed GenINR with FAF integration. Subretinal fluid (examples above and below left) is common in CSCR, whereas intraretinal fluid (above right) is seldom). The example below right shows a case with photoreceptor atrophy. In all cases, the retinal shape is reproduced well, but the images appear blurred in the areas of displaced retinal layers and no blood vessel shadows are visible. The color coding is the same as in Fig.~\ref{fig:qualitative_results_slo}.}
    \label{fig:qualitative_results_cscr}
\end{figure*}

\section{Experiments and Results}
\subsection{Anisotropic Interpolation}
The interpolation capabilities of the proposed INR are first assessed using the healthy dataset. To do so, the network is trained with 80 OCT volumes and SLO images of 40 volunteers using only 16 B-scans per volume as input to the INR, resulting in even higher inter-slice distances than for typical clinical images. The input B-scans are sampled equidistantly from the entire depth of the image volumes, enabling evaluation on intermediate B-scans. For comparison, we use simple linear interpolation, registration-based interpolation~\citep{ehrhardt2007structure} and non-generalizable INRs adapted to single OCT images. To segment intermediate B-scans for the interpolation methods, we proceed as follows: For the linear interpolation, we interpolate the label images linearly and round the resulting continuous values to integers. For the registration-based method, we use the deformation fields resulting from the B-scan interpolation and apply them to the label images using nearest neighbor interpolation. The non-generalizable INRs (SingleINRs) are implemented with a slightly changed architecture, since we observed improved performance for the modified architecture. The SingleINR, thus, consists of three linear layers of size 512 with sinusoidal activation functions~\citep{sitzmann2020implicit}. Furthermore, to get the SingleINR to interpolate between the given B-scans instead of overfitting to the given positions, we intentionally use smaller B-scan distances than the real ones so that the input coordinates lie approximately on an isotropic grid.

Since the proposed generalizable INR (GenINR) is the only one of the considered methods that can be used to interpolate and segment unseen test cases, we additionally evaluate the GenINR for the 20 unseen cases from the healthy dataset. For these cases, new latent priors are learned, while the INR weights remain frozen. We repeat this experiment for the pathological data, using 120 OCT and FAF images from 14 patients for training and 26 images from five patients for testing. Here, we use every second B-scan for training, i.e., 13 B-scans per volume, and evaluate performance for the remaining twelve B-scans. Results for the healthy dataset can be found in Tab.~\ref{tab:interp_results_healthy} and in Tab.~\ref{tab:interp_results_cscr} for the CSCR dataset. For both datasets, several image similarity metrics, including mean absolute error (MAE), peak signal-to-noise ratio (PSNR), structural similarity index measure (SSIM), and learned perceptual image patch similarity (LPIPS), are reported to assess the image reconstruction and interpolation capability of the considered methods. For a consistent evaluation of both proposed approaches, we used the redefined class labels in which the choroid is discarded and very thin layers are merged. To evaluate the segmentation performance, the Dice score averaged over the nine retinal layers, as well as the average symmetric surface distance (ASSD) and Hausdorff distance (HD) of the layer borders, are reported for the healthy dataset. For the CSCR dataset, we evaluate the segmentation performance with Dice scores of the retina ($\text{Dice}_\text{retina}$) and of the three disease-related fluid accumulations combined into one binary pathology label ($\text{Dice}_\text{fluid}$). 

The quantitative results for the healthy dataset show that our proposed GenINR achieves the best results for three out of four image similarity metrics. Only for the LPIPS metric, the GenINRs underperform, while the simple interpolation methods (linear and registration-based) achieve the highest values. Simple interpolation hardly changes the contrast and noise level, whereas the GenINR interpolation results look comparably smooth, missing the noisy and high-frequency components of the images (cf. Fig.~\ref{fig:qualitative_results_slo}), which might explain the LPIPS results. Regarding image segmentation, the proposed method performs better than the instance-based methods, indicating a better interpolation of the retinal shape achieved by generalization using population-based training. Especially in the fovea region, where large shape differences can be observed for neighboring B-scans, the GenINR is the only method that can meaningfully represent the shape, as shown in the examples on the left side in Fig.~\ref{fig:qualitative_results_slo}. The SingleINR, in turn, shows a strong overfitting on the slices used for training, hindering it from interpolating the retinal shape correctly. This overfitting also prevents the SingleINR from benefiting from the additional SLO information; in fact, the additional information, which is different at the interpolated positions than at those used for training, causes SingleINR to fail. The GenINR, in turn, is enabled to reconstruct blood vessels, as shown in the examples on the right in Fig.~\ref{fig:qualitative_results_slo}. The localization information provided by the en-face SLO enables the correct positioning of blood vessels and the reconstruction of blood vessel shadows in the outer retina. Reconstruction in the fovea region, where fewer blood vessels are located, appears to be more challenging, as no reconstructions of vessels or vessel shadows are notable. Interpolation artifacts from neighboring B-scans in the form of misplaced shadows, as observed in competing methods, do not occur for the GenINR -- all observed vessel shadows are correctly placed. Furthermore, the GenINR enables the interpolation and segmentation of unseen test cases, which is not possible with the instance-specific methods. Although a decline in performance for the test cases compared to the training images can be observed, the GenINR still manages to reconstruct the OCT volumes reliably and to represent the retinal layer boundaries with an average ASSD of $\SI{7.0}{\micro \metre}$. 

Similar results can be observed for the CSCR data. Again, the GenINR demonstrates better performance than the baseline methods both in terms of image reconstruction and segmentation, except for the LPIPS metric. As for the healthy data, the simple linear interpolation delivers the best results for this metric. As observed before, the retinal shape is best aligned with the generalizable INR (average Dice score of 98.98~\% for the retina), and also the pathologies are better interpolated. For the test data, all metrics considered show similar performance to the training data, except for the Dice score of pathologies. Without FAF integration, an average Dice score of 16.84~\% is achieved for the interpolated slices of the test images and improved slightly to 17.59~\% with the additional FAF modality. Although there is no known clear relationship between FAF and OCT appearance, the INR seems to benefit from the additional modality, indicating a direct correlation between FAF and OCT. The reduced segmentation performance for the test cases indicates that the highly variable appearance of retinal pathologies in CSCR cannot yet be completely covered by the individualization through the latent codes. Some exemplary interpolation results for the CSCR data are shown in Fig.~\ref{fig:qualitative_results_cscr}. Here, it can be seen that the retinal layers are reconstructed well for B-scans without severe pathological changes. However, no reconstruction of blood vessels can be observed, which we attribute to the distorting effect of pathologies on training. Also, in pathological areas the retinal layers appear more blurred compared to the regions unaffected by fluids. Still, the shape of both the retina and the pathologies is accurately reproduced by the GenINR$_\text{FAF}$.

Across all experiments, the runtime for each subject was below one second for linear interpolation, around 19 seconds for registration-based interpolation, approximately three minutes for the GenINR (inference adaptation to a new subject), and about five minutes for the SingleINR.

\subsection{Atlas Registration}\label{sec:atlasRegistration}

\begin{table*}[h]
  \centering
  \caption{Quantitative performance comparison of the atlas registration methodology. Similarity metrics, deformation metrics, and the inference time per subject are reported. GenINR$_{\text{init}}$ shows the initial alignment of the generated atlas without registration. Metrics are stated as the \meanstd{mean}{std}. The best results are highlighted.}
  \label{tab:RegResultTable}
  \begin{tabular}{@{}lllllll@{}}
  \toprule
  & \multicolumn{3}{l}{Similarity} & \multicolumn{2}{l}{Deformation} \\
  Method & $\downarrow$ $\text{ASSD}^{\mathtt{[\SI{}{\micro\metre}]}}$         & $\uparrow \text{Dice}^{\mathtt{[\%]}}$    & $\uparrow$ $\text{SSIM}^{\mathtt{[\%]}}$   & $\downarrow$ $\left\lvert J_\varphi \right\rvert \leq 0^{\mathtt{[\%]}}$  & $\downarrow \left\lVert u(x,y,z) \right\rVert_1$ & $\downarrow$ Time$^{[\SI{}{\second}]}$\\\midrule
  $\text{GenINR}_{\text{init}}$ & \meanstd{19.4}{5.8} & \meanstd{76.4}{7.9} & \meanstd{38.5}{4.6}  & -- & -- & -- \\  
  $\text{GenINR}_{\text{aff}}$ & \meanstd{14.3}{3.3} & \meanstd{82.9}{4.3}& \meanstd{41.7}{3.7} & -- & -- & -- \\
  SINA & \meanstd{29.8}{11.0} & \meanstd{65.1}{10.3} & \meanstd{35.1}{4.9} & \meanstd{\textbf{0.0}}{0.0} & \meanstd{0.009}{0.003} & \meanstd{279.3}{32.52} \\
  SingleINR & \meanstd{12.8}{2.3} & \meanstd{84.3}{2.9} & \meanstd{41.9}{4.0} & \meanstd{0.001}{0.007} & \meanstd{\textbf{0.005}}{ 0.003} & \meanstd{97.3}{4.6} \\
  $\text{GenINR}$ & \meanstd{\textbf{11.3}}{1.6} & \meanstd{\textbf{86.4}}{2.3} & \meanstd{\textbf{42.4}}{3.6} & \meanstd{\textbf{0.0}}{0.0} & \meanstd{0.04}{0.011} & \meanstd{\textbf{23.1}}{0.85} \\
  \bottomrule
  \end{tabular}
  \end{table*}

\begin{figure*}[h]
    \centering
    \includegraphics[width=\linewidth]{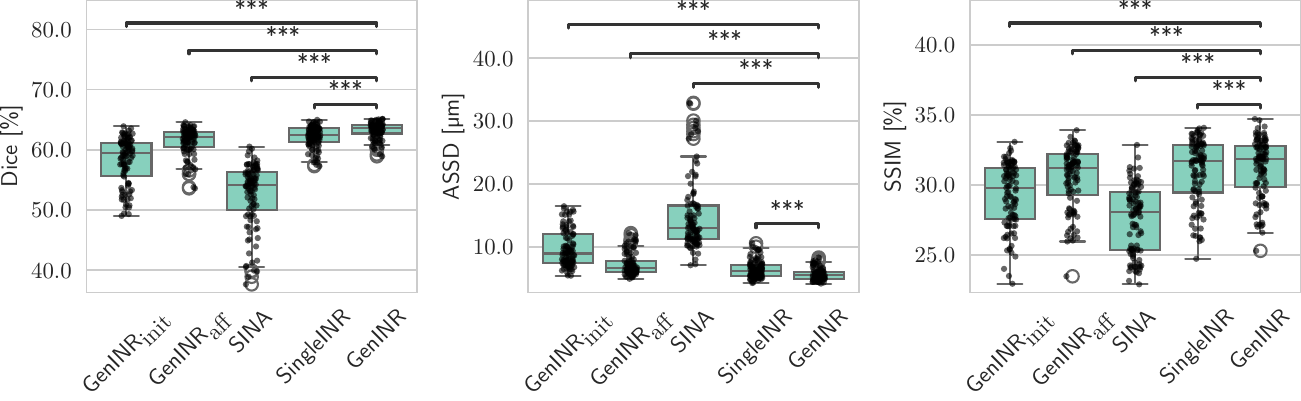}
    \caption{Comparison of the population-based atlas registration methods across all similarity metrics. Statistical significance is indicated using stars (***: $p < 0.001$), determined via the Wilcoxon signed-rank test and Holm-Bonferroni correction using the Python package by \cite{charlier_2022_plots}.}
    \label{fig:reg_sim_metrics}
\end{figure*}

This section assesses the usage of generalizable INRs for image registration tasks -- evaluating how effectively they capture structural consistency within a population. Registration performance and atlas generation capabilities are evaluated using the healthy dataset and compared with two different approaches. First, we compared our GenINR framework with the learnable implicit neural atlas approach of \cite{grossbrohmer2024sina} (SINA), which uses B-spline control points for registration. For the purpose of supervised segmentation, SINA is extended by a second INR that acts as a label atlas, which introduces weak supervision into the atlas generation process. In addition, we modified our proposed method with an non-generalizable INR (SingleINR). Since SingleINRs are trained on individual instances, joint optimization of atlas and deformation presents new challenges. Thus, we used the computed voxel-wise median as a fixed atlas that is registered using the inter-subject approach of \cite{wolterink2022implicit}. The SingleINR consists of a SIREN with four hidden linear layers of size 128 \citep{sitzmann2020implicit}. As a reference, we include the initial alignment to the atlas (GenINR$_{\text{init}}$), and the alignment after affine registration (GenINR$_{\text{aff}}$). 

Intensity similarity is evaluated using the SSIM, while anatomical alignment is evaluated using the Dice and the ASSD. Deformation's integrity and plausibility are evaluated using the percentage of negative Jacobian determinants ($\left\lvert J_\varphi \right\rvert \leq 0$) and the size of the displacements, as measured by the L1 norm ($\left\lVert u(x,y,z) \right\rVert_1$). All similarity metrics are evaluated on single B-scans and averaged over each volume. The methods are trained using the healthy dataset with 64 equidistantly sampled B-scans. The experiments are performed using five-fold cross-validation with 80 OCT and segmentation volumes for training and 20 volumes for inference. We conducted four experiments to evaluate the performance of our proposed registration framework. We 1) evaluate the atlas generation capabilities, 2) investigate the registration performance, 3) show that the framework can be used independently of the inference resolution, and 4) perform an ablation study by replacing the implicit atlas representation with an explicit discrete parameter representation.

\begin{figure*}[h]
    \centering
    \includegraphics[width=\linewidth]{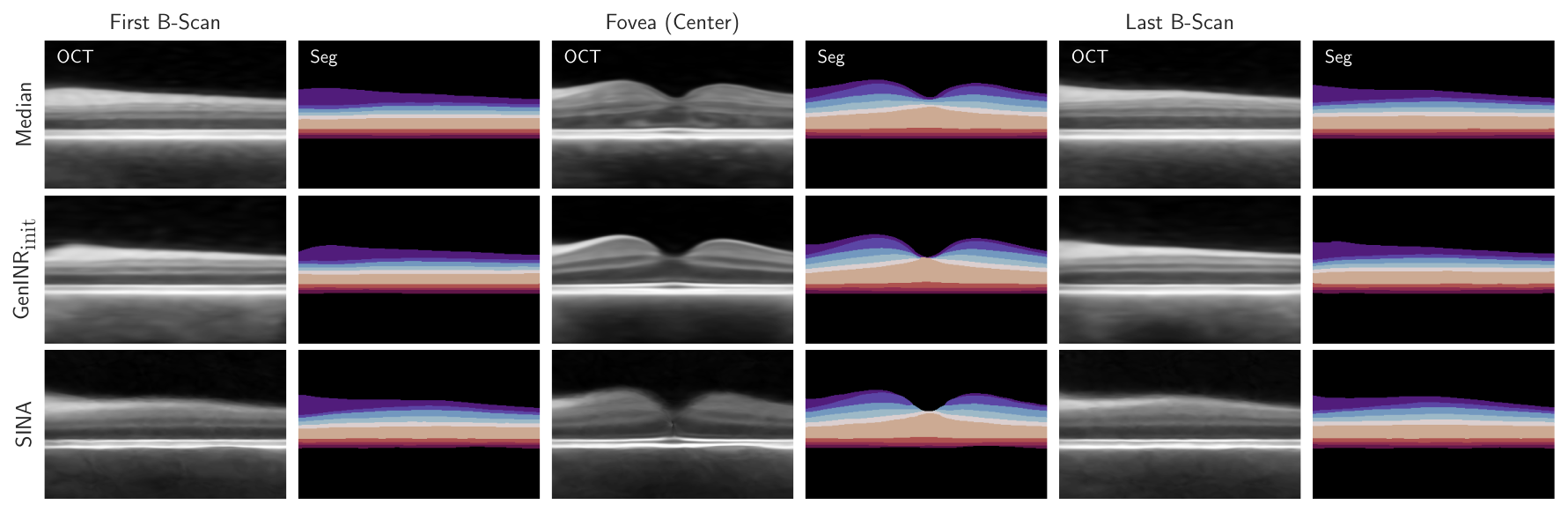} 
    \caption{Three examples of B-scans located at the center and each side of the volume. Compared are the voxel-wise median, our learned atlas, and the learned atlas of SINA. For each B-scan, the intensity values and the segmentation map are displayed. }
    \label{fig:reg_qualitativ_results}
\end{figure*}

Our proposed registration framework creates a well-aligned mean representation, i.e. atlas, as indicated by GenINR$_{\text{init}}$ in Tab.~\ref{tab:RegResultTable}. It achieves scores of $\SI{19.4}{\micro \metre}$, $76.4\, \%$, and $38.5 \; \%$ for ASSD, Dice, and SSIM, respectively. These findings are supported by the qualitative results shown in Fig.~\ref{fig:reg_qualitativ_results}. We consider an atlas to be good when it is sharp -- especially on the layer edges, artifact-free, and anatomically plausible. Our framework shows qualitatively sharper and more precise layer separation as well as hardly any artifacts compared to the median representation or SINA. Moreover, SINA exhibits artifacts at the center of the fovea and, in general, a blurrier representation. While the median atlas already shows minimal artifacts and reasonable layer separation, our method achieves higher sharpness and better separation in the upper retinal layers. Although both generated atlases show a consistent alignment between class labels and intensity representations, our method yields more anatomically plausible results, particularly in the innermost retinal layers, with a continuous representation of the RNFL across the retina.

\begin{figure*}[h]
    \centering
    \includegraphics[width=\linewidth]{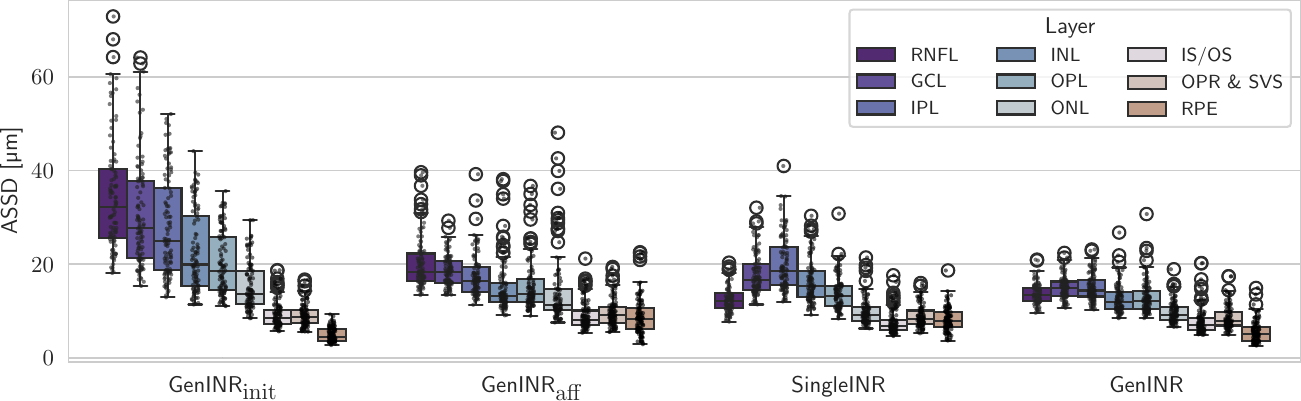}
    \caption{Detailed performance overview of the SingleINR and GenINR on the example of ASSD for each retinal layer: retinal nerve fiber layer (RNFL), ganglion cell layer (GCL), inner plexiform layer (IPL), inner nuclear layer (INL), outer plexiform layer (OPL), outer nuclear layer (ONL), IS/OS junction (IS/OS), outer photoreceptor segment (OPR), subretinal virtual space (SVS), and retinal pigment epithelium (RPE).}
    \label{fig:reg_assd_detail}
\end{figure*}

Focusing on the results of the atlas-to-subject registration experiment in Tab.~\ref{tab:RegResultTable}, our approach achieved the best results by about $\SI{11.3}{\micro\metre}$, $86.4 \,\% $, and $42.4 \, \%$ for ASSD, Dice, and SSIM, respectively. Significant improvements are shown in Fig.~\ref{fig:reg_sim_metrics} for GenINR over the initial alignment, affine alignment, and the two approaches that were compared: SINA and SingleINR. Meanwhile, the time taken for inference execution is reduced by around four minutes per subject compared to SINA and around one minute per subject compared to the SingleINR -- resulting in faster processing at around $\SI{23.1}{\second}$ per subject. However, SINA presents a significant deterioration in the segmentation metrics compared to the base alignment (GenINR$_{\text{init}}$), indicating an insufficient number of control points of the interpolated displacement field. SINA and GenINR produce smooth and fold-free deformations without the need for additional regularization. SingleINR exhibits a slight percentage of folding. Fig.~\ref{fig:reg_assd_detail} presents the anatomical alignment results using the separated ASSD metric for all layers. Due to pre-processing, the bottom layers (RPE, OPR \& SVS, and IS/OS) are already well aligned initially and challenge the methods to generate a proper transition from small to larger displacements. Despite this, our framework achieved a uniform performance distributed across all layers, whereas the SingleINR struggled with the upper middle layers GCL, IPL, and INL.

\begin{figure*}[h]
    \centering
    \includegraphics[width=\linewidth]{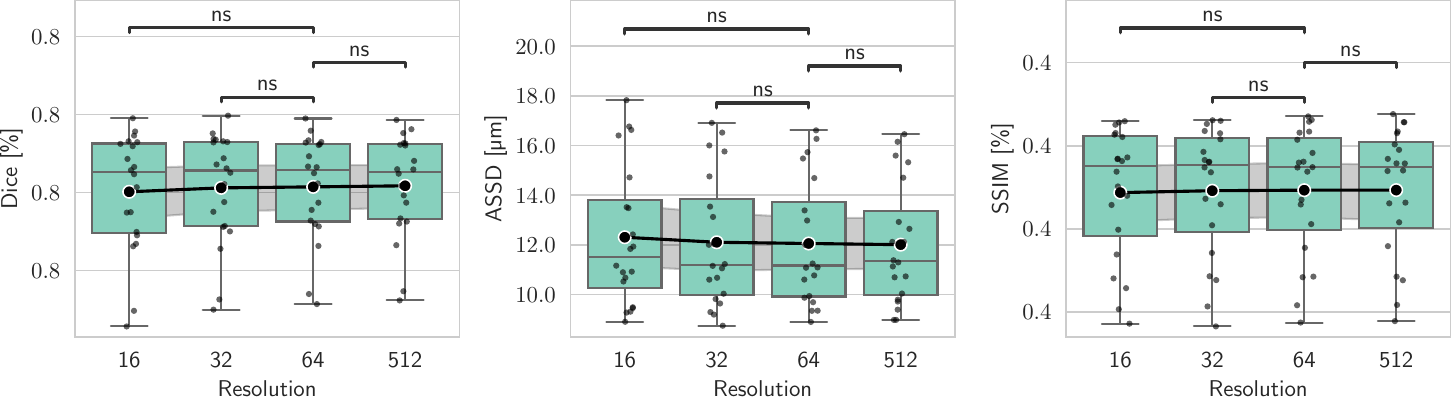}
    \caption{Results of our GenINR during inference -- trained and evaluated on different numbers of B-scans. No statistical significance is indicated using (ns: not significant, $p \geq  0.01$), determined via the Wilcoxon signed-rank test and Holm-Bonferroni correction using the Python package by \cite{charlier_2022_plots}.}
    \label{fig:reg_metrics_res}
\end{figure*}

Due to the continuous nature of the INR and generalizable INR, the proposed framework is independent of the discrete grid representation and the associated resolution. Therefore, during inference, the latent codes of the trained atlas framework are learned and evaluated on 16, 32, 64, and 512 B-scans. Fig.~\ref{fig:reg_metrics_res} illustrates no significant changes ($p> 0.01$) in the alignment performance when compared with the resolution of 64 B-scans and, in general, similar performance scores. All setups result in around 85.2$\,\%$, $\SI{12.1}{\micro \metre}$, and 42$\,\%$ for Dice, ASSD, and SSIM, respectively.

We conducted an ablation study to illustrate the performance gains of the implicit atlas representation. To do so, we replaced the INR with two learnable parameter images for intensities and class labels, which are then transformed using the deformation field and linear interpolation to be compared with the target volume. This explicit atlas representation scales $\mathcal{O}(n^3)$ with the size of the three dimensions $n$, which limited the experiment to a resolution of 16 B-scans. The results are shown in Fig.~\ref{tab:ResultsRegAblationStudy} and support the advantage of implicit atlas representation by even improving alignment performance across all segmentation-based metrics. It should be noted that the training time refers to the latent prior training during inference. 

\begin{table}[h]
  \centering
  \caption{Results of the ablation study conducted on 16 B-scans. Similarity metrics, deformation metrics, and the inference time per subject are reported. Metrics are stated as the \meanstd{mean}{std}. The best results are highlighted.}
  \label{tab:ResultsRegAblationStudy}
  \begin{tabular}{@{}lllll@{}}
  \toprule
  Atlas & $\downarrow$ $\text{ASSD}^{\mathtt{[\SI{}{\micro\metre}]}}$         & $\uparrow \text{Dice}^{\mathtt{[\%]}}$    & $\uparrow$ $\text{SSIM}^{\mathtt{[\%]}} $ &   $\uparrow$ Time$^{[\SI{}{\second}]}$\\\midrule
 Explicit & \meanstd{11.2}{1.8} & \meanstd{86.1}{2.6} & \meanstd{\textbf{42.7}}{3.9} & \meanstd{15.5}{0.63} \\
 Implicit & \meanstd{\textbf{10.9}}{2.1} & \meanstd{\textbf{86.7}}{3.1} & \meanstd{41.8}{4.7} & \meanstd{\textbf{10.6}}{0.85} \\
  \bottomrule
  \end{tabular}
  \end{table}

\begin{figure*}[h]
    \centering
    \includegraphics[width=\textwidth]{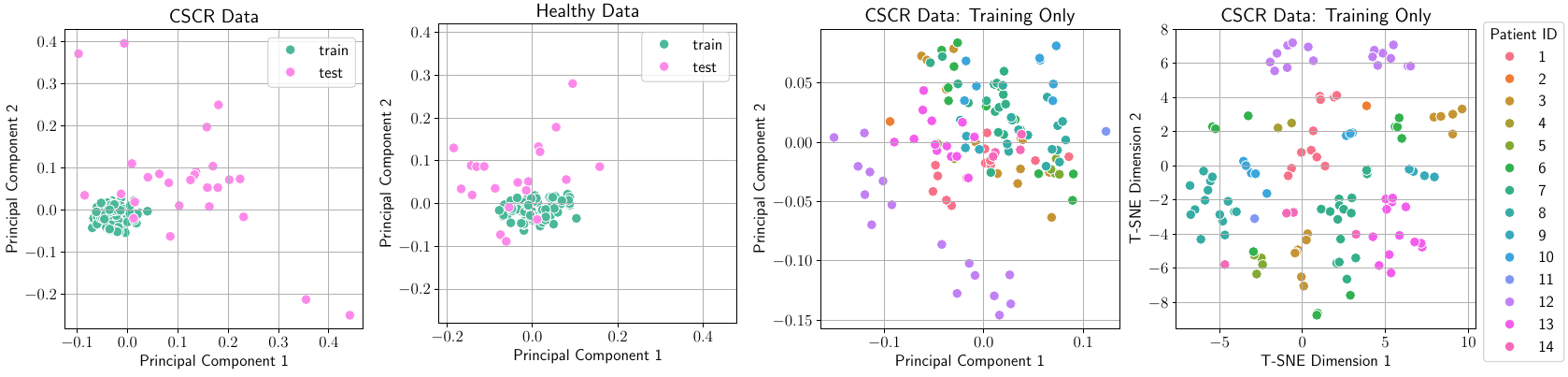}
    \caption{PCA analysis of the learned latent codes of the interpolation INR (GenINR$_\text{SLO}$ for the healthy data and GenINR$_\text{FAF}$ for the CSCR data). On the left-hand side, the PCA for the latent codes of training and test cases is performed together. The two plots on the right show a PCA and t-SNE performed exclusively on the training images of GenINR$_\text{FAF}$ and are color-coded by subject.}
    \label{fig:interp-pca}
\end{figure*}

\subsection{Latent Space Analysis}
To gain a better insight into the INRs, we perform a principal component analysis (PCA) on the latent priors for both approaches.
Fig.~\ref{fig:interp-pca} shows the result of the PCA analysis of the first two principal components for the inter-B-scan interpolation -- explaining about 20~\% of the variance -- as well as a t-SNE visualization of the training latent space. While the latent priors for the training data are more compactly centered around 0, the latent priors for the test cases deviate quite heavily from the training cases, showing that individual solutions are found for the test data rather than the most similar training case. The deviation is even stronger for the CSCR data, possibly due to a stronger variability in the data. Furthermore, training cases of the same patient appear to have grouped latent priors, indicating that individual structural differences are coded efficiently in the vector.

\begin{figure*}[h]
    \centering
    \includegraphics[width=\linewidth]{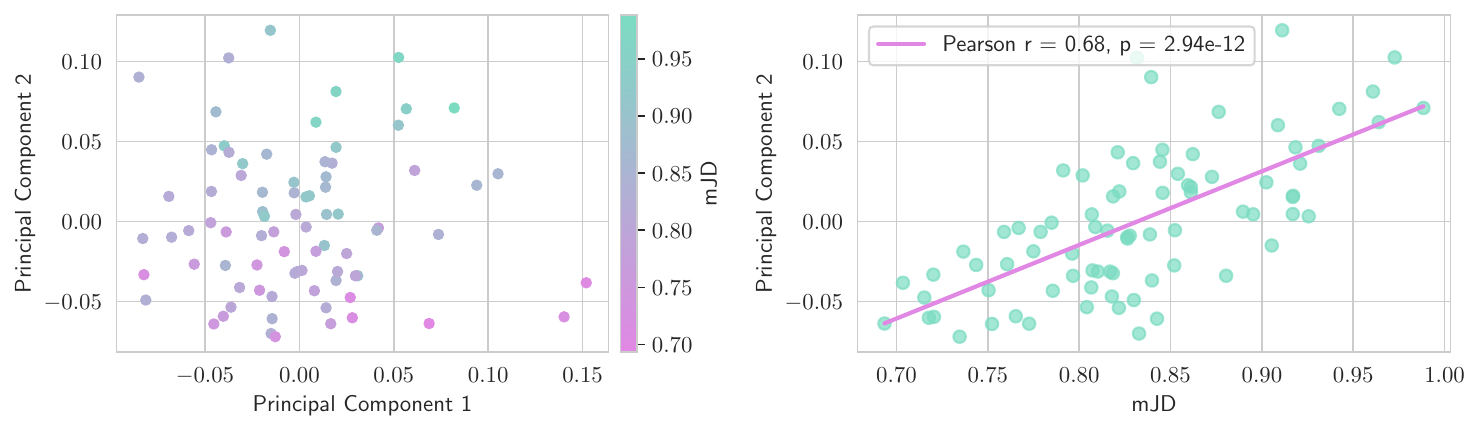}
    \caption{Exploration of the latent space using PCA and the masked mean of the Jacobian Determinant (mJD) as a metric to describe the general volume change. On the left side, the first two principal components are displayed with the mJD color-coded. On the right, the correlation between mJD and the second principal component is plotted. }
    \label{fig:RegPCA}
\end{figure*}

For the registration approach, we analyzed the latent priors and their relation to the deformation. We again performed a PCA on the latent priors and related the first two principal components (explaining around 13~\% of the variance) with the mean Jacobian determinant that is computed using the retina as a mask. Fig.~\ref{fig:RegPCA} shows the results, indicating a significant correlation between the mean Jacobian determinant -- which quantifies the volume changes caused by the deformation -- and the second principal component of the latent space.  

\section{Discussion}
In this work, we examined the usage of INRs for highly anisotropic OCT data. By representing the OCT volume as a continuous function, we obtain a volume of arbitrary resolution. Our results prove that the reconstructed space between B-scans carries meaningful information. This way, we overcome the limitation of established approaches.

Our proposed interpolation method achieves better performance than the considered competitive approaches. For healthy cases, it can interpolate the retinal layer shape most accurately while using even larger slice distances than clinical OCT volumes. On pathological data, we were able to repeat good OCT B-scan interpolation performance while retaining pathologies within these interpolated B-scans. Our approach is mainly limited by the low-frequency reconstruction by the generalizable INR, leaving a disparity in intensity reconstruction. Despite that, the generalizable INR enables the incorporation of en-face modality information and segmentation prediction of unseen OCT volumes. This in turn results in better reconstruction of retina shape, vessel structures, and fluid positions in interpolated slices. Further improvement might be gained by incorporating sub-pixel-level layer boundaries instead of the voxel-wise segmentation used in this work. Shape-based interpolation methods similarly make use of segmentations and boundaries and are well-suited for interpolating label maps, but do not address the B-scan interpolation task because intensity information is not modeled. In contrast, our approach generates supersampled OCT data with corresponding labels, which can benefit various downstream applications, such as volumetric segmentation, 3D reconstruction, longitudinal analyses or image registration, as in Sec.~\ref{sec:atlasRegistration}. Demonstrating direct clinical benefit remains an important direction for future work and will be addressed in subsequent studies.

To provide further analysis and understanding of the population, we built upon the previous findings and proposed a similar framework for registration and atlas generation. We combine methods from \cite{grossbrohmer2024sina} and \cite{wolterink2022implicit} with a generalization approach to obtain a pipeline that enables the joint prediction of a resolution-agnostic atlas and corresponding deformation fields. \cite{grossbrohmer2024sina} proposed an INR as an atlas representation but used a simple registration based on B-spline control points. This instance-based B-spline registration showed good performance on other datasets but -- as shown in this study -- scales poorly to the challenges associated with OCT. Implicit deformable image registration~\citep{wolterink2022implicit} on the other hand, exhibited good alignment performance with close to no foldings on our OCT data. To resolve the restriction of the SingleINR to prior knowledge, we proposed a generalizable INR registration framework to include population-based knowledge and auxiliary information (i.e., retinal layer segmentation) in a joint atlas and deformation optimization process. Thus, we enable the creation of a sharp and accurate atlas representation as well as a continuous representation of the deformation. The resulting method is, by construction, resolution-agnostic and can be used on volumes containing varying numbers of B-scans. Our combined framework improved the alignment and representation performance, reduced the execution time during inference, and reduced the deformation field representation to a single lower-dimensional vector. In line with the findings of \cite{tian2024nephi}, we show that generalizable INRs present an efficient approach to model the registration process. Since smoothness of the deformation field is intended, the drawback of low-frequency reconstruction observed for interpolation does not transfer to registration.

Altogether, INRs simplify the handling and processing of high-dimensional image data. Especially for image registration, INRs seem to cause a trend for new approaches \citep{wolterink2022implicit, sideri2024sinr, van2023robust, tian2024nephi, grossbrohmer2024sina}. When the image is represented by an INR, it eliminates the need for linear interpolation in the registration process while maintaining differentiability. Furthermore, implicitly representing data as functions reduces memory and runtime requirements when compared to other iterative approaches. This is particularly the case once the INR is combined with a generalization approach and the representation is compressed into the latent priors. With this, new potentials and analysis options arise, as we demonstrated for our proposed methods.


\section{Conclusion}
This work explored the potential of INRs in a highly an\-i\-so\-tro\-pic domain such as retinal OCT imaging. We proposed two methods for interpolation and atlas registration using generalizable INRs. Our findings highlight the power of INRs as a resolution-agnostic interpolation or registration tool. With both approaches, we were able to generate more accurate reconstructions than the considered baselines. Our approach overcomes limitations regarding the resolution introduced by device-specific sampling and enables the fusion of different modalities.

Generalizable INRs offer the advantage of low-di\-men\-sio\-nal data representations due to latent priors. These latent vectors provide a more efficient way to store the data and also enable the analysis of the underlying structures. 
That way, we visualized relations in the latent space -- suggesting new use cases such as classification or anomaly detection. In addition, further modalities such as OCT-angiography or fluorescence lifetime imaging ophthalmoscopy may be evaluated. 
In summary, INRs, and generalizable INRs in particular, are a powerful tool for medical image analysis. They highlight the value of continuous representation by enabling a resolution-agnostic interpolation and atlas registration framework. 

\acks{This work was supported by funding from the federal state of Schleswig-Holstein, Germany (grant 22024003).}
\newpage

%
\ethics{The work follows appropriate ethical standards in conducting research and writing the manuscript, following all applicable laws and regulations regarding treatment of animals or human subjects.}

\coi{We declare we don't have conflicts of interest.}

\data{
The CSCR dataset was acquired during clinical routine and is not publicly available due to data protection regulations.
The in-house dataset of healthy volunteers is currently being prepared for public release. Further details will be provided in the \href{https://github.com/tkepp/ResA-OCT}{GitHub repository} after release.

}

\bibliography{sample}


\end{document}